\pdfoutput=1

\documentclass[11pt]{article}

 \usepackage[final]{acl}

\usepackage{times}
\usepackage{latexsym}
\usepackage[T1]{fontenc}
\usepackage[utf8]{inputenc}
\usepackage{microtype}
\usepackage{inconsolata}
\usepackage{graphicx}
\usepackage{booktabs} 
\usepackage{multirow}
\usepackage{tabularx}
\usepackage{makecell}
\usepackage{array}
\usepackage{float}
\usepackage{enumitem}
\usepackage{lipsum}
\usepackage{hyperref}
\usepackage[utf8]{inputenc}
\usepackage{graphicx}
\usepackage{fancyvrb}
\usepackage{framed}
\usepackage{multicol}
\usepackage{comment}
\usepackage{amsmath}
\usepackage{stfloats}

\title{Learning to Verify Summary Facts with Fine-Grained LLM Feedback}

\author{Jihwan Oh,
Jeonghwan Choi, Nicole Hee-Yeon Kim, Taewon Yun,
Hwanjun Song\thanks{Corresponding Author.}\\
Korea Advanced Institute of Science and Technology\\
\texttt{\{jh.oh, hwani.choi, nicolekim, ytaewon0415, songhwanjun\}@kaist.ac.kr}}
\begin{document}

\maketitle

\begin{abstract}
Training automatic summary fact verifiers often faces the challenge of a lack of human-labeled data. In this paper, we explore alternative way of leveraging Large Language Model (LLM) generated feedback to address the inherent limitation of using human-labeled data. 
We introduce FineSumFact, a large-scale dataset containing fine-grained factual feedback on summaries. We employ 10 distinct LLMs for diverse summary generation and Llama-3-70B-Instruct for feedback. We utilize this dataset to fine-tune the lightweight open-source model Llama-3-8B-Instruct, optimizing resource efficiency while maintaining high performance. Our experimental results reveal that the model trained on extensive LLM-generated datasets surpasses that trained on smaller human-annotated datasets when evaluated using human-generated test sets. Fine-tuning fact verification models with LLM feedback can be more effective and cost-efficient than using human feedback. The dataset is available at \href{https://github.com/DISL-Lab/FineSumFact}{https://github.com/DISL-Lab/FineSumFact}

\end{abstract}

\begin{figure}[ht]
    \centering
    \includegraphics[width=\linewidth]{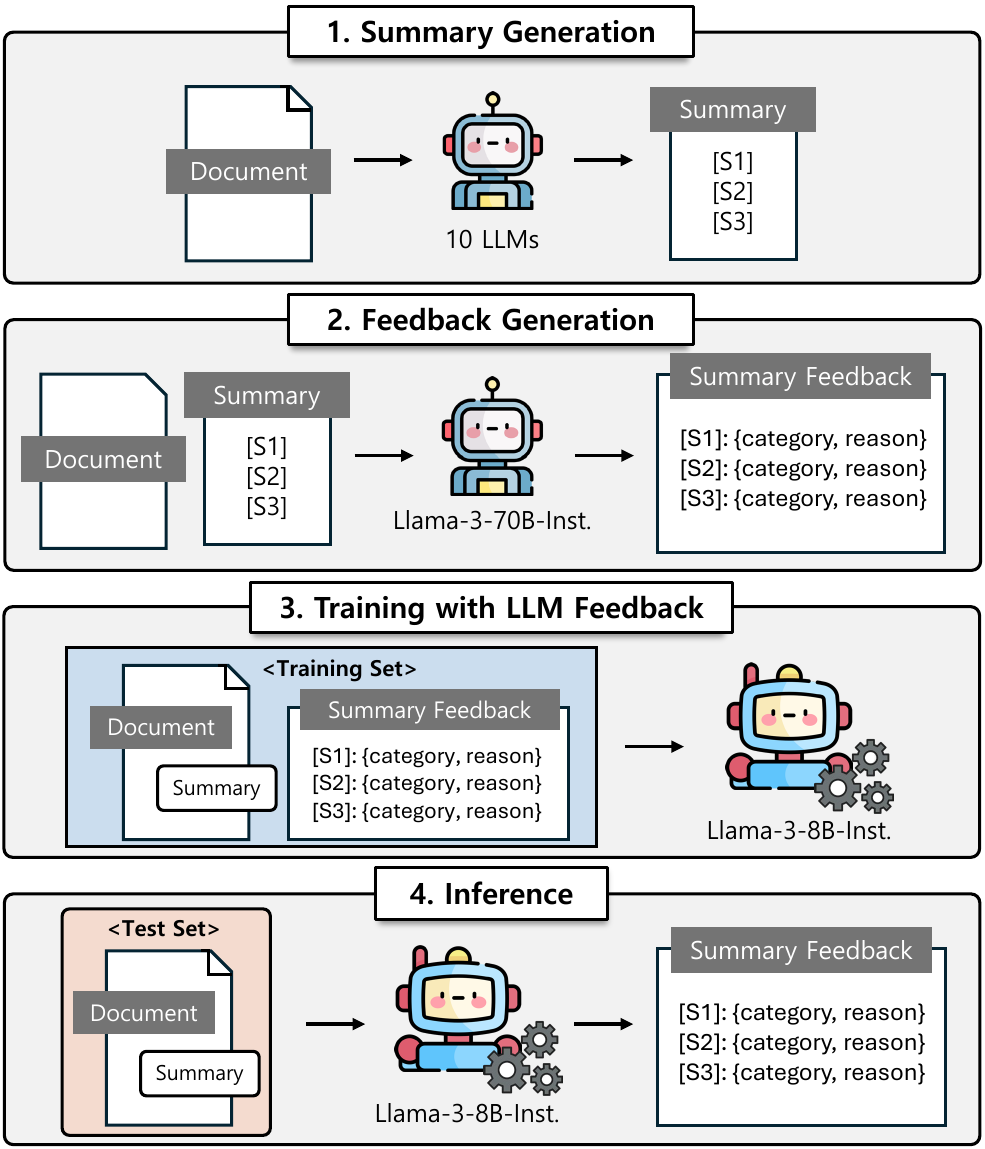} % 그림의 너비를 텍스트 너비에 맞춤
    \vspace*{-0.75cm}
    \caption{{Pipeline}: our evaluator is trained with LLM feedback generated on diverse input texts and summaries and then tested on an unseen test set.}
    \label{fig:example}
    \vspace*{-0.6cm}
\end{figure}

\section{Introduction} 

%\footnotetext[1]{NON-ARCHIVAL submission.}
% \footnote{\url{https://openai.com/index/gpt-4-research/}}
%\footnote{\url{https://ai.meta.com/blog/meta-llama-3-1/}} 
Recent advancements in large language models (LLMs) have significantly enhanced the text summarization performance\,\citep{tang2024tofueval, zhang2024benchmarking}. State-of-the-art models such as GPT-4 excel at generating coherent summaries from extensive datasets, processing input contexts exceeding 100k tokens, thereby significantly enhancing their summarization capabilities\,\cite{ravaut2023context}. However, hallucination issues still occur in summaries, highlighting the importance of summary fact verification\,\citep{cao2022hallucinated}.

Verifying the fact of the summaries inevitably necessitates considerable human effort, rendering the evaluation process both time-intensive and cost-prohibitive. In manual evaluation, non-expert human evaluators are often tasked with labeling summaries across diverse domains \cite{geiger2020garbage}. 
%whether through crowdsourcing or expert engagement
In particular, this process gets more costly and challenging to reproduce at a fine-grained level evaluation, such as error localization and explainable evaluation.
%Labeling doc-summary pairs with human annotations, particularly at a fine-grained sentence level, is more costly and challenging to reproduce.
%\\
%

To mitigate the human cost involved, an alternative way is to employ AI-assisted labeling approaches\,\cite{desmond2021increasing, wang2021want} and the training of language models using LLM-generated labels, also known as knowledge distillation \cite{pangakis2024knowledge}. However, the application of knowledge distillation for fact verification remains unexplored. % csanády2024llambertlargescalelowcostdata

In this paper, we unveil the potential of using LLM-generated fine-grained feedback to train an efficient and effective fact verification model. As shown in Figure \ref{fig:example}, our pipeline consists of four stages: \textbf{(1) Summary Generation}: we generate diverse summaries using 10 different language models on collected input documents, which span from short to lengthy texts from non-dialogue to dialogue sampled from 7 distinct data domains; \textbf{(2) Feedback Generation}: we acquire a large volume of fine-grained LLM feedback using an off-the-shelf evaluator, FineSurE (using Llama-3-70B-Instruct)\,\cite{song2024finesure}, producing sentence-level fact verification labels along with error types; \textbf{(3) Training with LLM Feedback}: we fine-tune a much smaller Llama-3-8B-Instruct model with LLM feedback through sequence-level knowledge distillation\,\cite{kim2016sequence}, leading to an efficient automated verifier; and \textbf{(4) Inference}: we evaluate the distilled model on unseen document-summary pairs to check the agreement with human judgments.

%we propose to utilize LLM-generated labels to train a automatic summary evaluation model. We collected documents of varying lengths from 5 to 4,000 tokens, from various domains and generated numerous summaries using 10 different LLMs. Utilizing Llama-3-70b, we then generated sentence-level labels to assess the factual consistency of each summary with its corresponding document, considering reason with 9 different error categories.

Our key findings are: (1) Training with a large amount of LLM-generated feedback can outperform using a limited set of human feedback in automated evaluation; (2) Evaluation accuracy improves considerably when trained with explainable feedback (e.g., reasoning, error types); and (3) Increasing the volume of training data with LLM feedback correlates positively with enhanced model performance.

\vspace*{-0.1cm}
\section{Related Work}
\vspace*{-0.1cm}

\paragraph{Fact Verification Datasets.}
Several datasets collected human annotations for training summary fact verification models. REALSumm \cite{bhandari2020re} provides a rigorous evaluation of 25 different summarizers, incorporating detailed human evaluations. SummEval \cite{fabbri2021summeval} offers a comprehensive benchmark with human annotations from both crowdsource and expert annotators.
In an effort to increase the scale of benchmark datasets, prior works, such as AggreFact \cite{tang2022understanding} and SummaC \cite{laban2022summac}, aggregated many human annotations from the previous benchmark datasets along with unified annotation schemes, focusing solely on the news domain. A separate line of research proposes a more fine-grained annotation framework. FRANK \cite{pagnoni2021understanding} introduces sentence-level feedback by categorizing factual errors into seven distinct types within the news domain, while TofuEval \cite{tang2024tofueval} proposes a complementary error taxonomy tailored to the dialogue domain, also providing sentence-level feedback.

% --- Before editing ----
% Several datasets have been proposed to train summary fact verification models. REALSumm \cite{bhandari2020re} provides a rigorous evaluation of 25 different summarization systems, incorporating detailed human evaluations. SummEval \cite{fabbri2021summeval} offers a comprehensive benchmark with human annotations from both crowdsource and expert annotators.
% FRANK \cite{pagnoni2021understanding} addresses the challenge of assessing factual consistency but potentially inaccurate summaries. 
% SummaC \cite{laban2022summac} leverages natural language inference (NLI) techniques to detect inconsistencies. 
% AggreFact \cite{tang2022understanding} aggregates factuality error annotations from nine existing datasets, categorizing them based on the summarization models used in each dataset.
% %SummEdits \cite{laban2023summedits} evaluates the ability of LLMs to reason about facts and to detect inconsistencies across 10 diverse domains. 
% TofuEval \cite{tang2024tofueval} targets the evaluation of factual consistency on topic-focused dialogue.

\paragraph{Fact Verification Methods.}

Various methods and metrics have been studied to verify factual consistency between documents and their summaries. FalseSum\,\cite{utama2022falsesum} generates document-level Natural Language Inference (NLI) examples with intentional factual inconsistencies to train evaluator models. QAFactEval\,\cite{fabbri2022qafacteval} is a QA-based metric, extracting information units from summaries and generating questions based on these units. Most recently, a few works simply rely on zero-shot inference, such as G-Eval\,\cite{liu2023improving} and FineSurE\,\cite{song2024finesure}.

Unlike prior studies, we construct a dataset without human effort by employing a recent LLM-based evaluator. We then train a lightweight open-source model, addressing open questions on the effectiveness of using LLM-based fine-grained feedback for fact verification.

\section{Preliminary}
\label{sec:preliminary}

\paragraph{Dataset with Human Feedback.} Datasets with human fact labels are widely used to train and test automated fact verifiers. For a more complete evaluation, we aggregate all the available human-labeled datasets for sentence-level fact verification, including AggreFact~\cite{tang2022understanding}, DiaSumFact~\cite{zhu2023annotating}, TofuEval\,\cite{tang2024tofueval}, and Ramprasad’24\,\cite{ramprasad2024evaluating}.
The aggregated data contains 6,546 document-summary pairs, each of which has sentence-level binary labels -- ``0" for no error and ``1" for a fact error. 85\% of pairs are used for training a fact verifier (one of our baselines) and the remaining 15\% of those are used for testing all the compared verifiers.
See the details in Appendix \ref{sec:human-labeled-data}. \looseness=-1

%We aggregate human-labeled data at the sentence level from existing studies, which differ in terms of granularity and annotation schemes when labeling factual consistency. 
%pair to a multi-class label \(y \in \{0, 1, 2, \ldots, C\}\), where 0 represents "factual" and 1 through C representes C types of "non-factual" categories. To create a unified dataset, we converted these annotations into a binary classification scheme \{0, 1\} where 0 represents "factual and 1 represents "non-factual". We include details on metadata in the Appendix.
\section{Learning with LLM Feedback}

\begin{table*}
\centering
\begin{small}
\newcolumntype{Y}{>{\centering\arraybackslash}X}
\begin{tabularx}{\textwidth}{l| l| Y Y Y Y}
\toprule
\multirow{2}{*}{Type} & \multirow{2}{*}{Method} & Sentence-level & Summary-level & System-level \\
\cmidrule(lr){3-3} \cmidrule(lr){4-4} \cmidrule(lr){5-5}
& & bAcc & Pearson Corr & Rank Corr \\
\midrule
QA-based (w. fine-tuned) & QAFactEval & - &  0.506 (0.000) & 0.864 (0.000) \\
NLI-based (w. fine-tuned) & SummaC-Conv & - & 0.337 (0.000) & 0.811 (0.000) \\\midrule
LLM-based (wo. fine-tuned) & Zero-shot & 57.4\% & 0.246 (0.000) & 0.663 (0.000) \\
\midrule
{LLM-based (w. fine-tuned)} 
& Human Feedback & 69.8\% & 0.534 (0.000) & 0.684 (0.000) \\
\cmidrule(lr){2-5}
& \textbf{LLM Feedback (Ours)} & \textbf{73.4}\% & \textbf{0.625} (0.000) & \textbf{0.865} (0.000) \\
\bottomrule
%ChatGPT-3.5-Turbo & zero-shot & 64.0\% & 0.385 (0.000) & 0.551 (0.004)\\
\end{tabularx}
\end{small}
\vspace*{-0.3cm}
\caption{Agreement with human judgments in fact verification on test data, where the value in the parenthesis is the p-value. All the LLM-based methods use Llama-3-8B-Instruct as the backbone. QAFactEval and SummaC do not support sentence-level fact verification. Further analysis on inference speed can be found in \ref{sec:latency-analysis}.}
% \label{tab:llm-performance-comparison}
\vspace*{-0.5cm}
\label{tab:main-result-exp}
\end{table*}

We build a large-scale dataset with LLM feedback to train a fact verifier capable of generalizing across various input contexts. Our dataset contains 10,877 documents, encompassing multiple domains, varying lengths, and two types (i.e., non-dialogue, dialogue). 
Particularly, the domains represented in the dataset include \emph{news} (CNN/DM: \citealt{hermann2015teaching}), \emph{interview} (MediaSum: \citealt{zhu2021mediasum}), \emph{daily} (DialogSum: \citealt{chen2021dialogsum}), \emph{meeting} (MeetingBank: \citealt{hu2023meetingbank}), \emph{knowledge} (WikiHow: \citealt{koupaee2018wikihow}), \emph{report} (GovReport: \citealt{huang2021efficient}), and \emph{medicine} (PubMed: \citealt{cohan2018discourse}). Refer to Appendix \ref{sec:llm-feedback-data-details} for detailed statistics and analysis. 

These source documents are used to construct labeled data with LLM feedback to train our fact verification model, following these three steps:

%The LLM-labeled dataset spans a broad spectrum of domains and accommodates documents of varying lengths. It includes 10,877 unique documents and 102,640 summaries, with an average document length of 890 words and an average summary length of 75 words. The domains represented in the dataset include \textbf{News} (CNN/DM: \citealt{NIPS2015_afdec700}), \textbf{Interview} (MediaSum: \citealt{zhu-etal-2021-mediasum}), \textbf{Daily} (DialogSum: \citealt{chen-etal-2021-dialogsum}), \textbf{Meeting} (MeetingBank: \citealt{hu-etal-2023-meetingbank}), \textbf{Knowledge} (WikiHow: \citealt{koupaee2018wikihowlargescaletext}), \textbf{Report} (GovReport: \citealt{huang-etal-2021-efficient}), and \textbf{Medicine} (PubMed: \citealt{cohan-etal-2018-discourse}).

\smallskip
\noindent\textbf{(1) Summary Generation:}
We generate summaries using 10 different LLMs to ensure a diverse distribution of summaries that include various types of fact errors. The summaries are generated by non-LLMs (BART-large-cnn, FLAN-T5-large, Pegasus-Large), open-source LLMs (Phi-2, Llama-2-13B-chat, Mistral-7B-Instruct, Mixtral-7B-Instruct), and commercial LLMs (Claude-Instant, GPT-3.5-turbo, GPT-4-turbo). 

\smallskip
\noindent\textbf{(2) Feedback Generation.} Ensuring high-quality feedback for fact-checking is essential. Hence, we adopt an off-the-shelf LLM-based fact verifier, FineSurE \citep{song2024finesure}, which produces fact error types and provides reasoning for the decisions. We use Llama-3-70B-Instruct as the backbone of FineSure since it exhibited the best-balanced accuracy of $92.0\%$ in the sentence-level fact check. The prompt of FineSurE is detailed in Figure \ref{fig:finesure-prompt} of Appendix \ref{sec:FineSurE_Prompts}. \looseness=-1
 As illustrated in Figure \ref{fig:example}, we acquire the feedback on nine fact error categories along with the reasoning behind the decision, including ``no error" (NoE), ``out of context error" (OutE), ``entity error" (EntE), ``predicate error" (PredE), ``circumstantial error" (CirE), ``grammatical error" (GramE), ``linking error" (LinkE), ``corefernce error" (CorefE), and ``other error" (See Appendix \ref{sec:error_types} for the error taxonomy). As a result, we collect LLM feedback on 102,640 document-summary pairs as the training data.

%Then, we generate the feedback for each summary at the sentence-level using Llama-3-70B-Instruct. Specific error types are identified for a detailed analysis using the FineSurE \citep{song2024finesure} evaluation framework.
%The sentence-level LLM-generated labels consist of a \textbf{binary label} indicating factual consistency or inconsistency, a \textbf{reason} explaining the reason for binary label, and an \textbf{error category} identifying 9 types of factual errors: No error, Out of context error (OutE), Entity error (EntE) Predicate error (PredE), Circumstantial error (CirE), Grammatical error (GramE), Linking error (LinkE), Corefernce error (CorefE), and Other error.

\smallskip
\noindent\textbf{(3) Training with LLM Feedback.}
We use QLoRA \cite{dettmers2024qlora} to fine-tune Llama-3-8B-Instruct on our training dataset with LLM feedback. We set the user prompt to be the same as FineSurE (in Figure \ref{fig:finesure-prompt}) and then set the assistant prompt to be the JSON output: {\sc [\{"sentence": "summary sentence 1", "reasoning": "reason", "category": "error type"\}, ...]}, which is the LLM feedback we obtained from FineSurE. We fine-tune the model for 8,000 iterations with a batch size of 32 using 4 NVIDIA H100 GPUs. By doing so, at inference time, we can parse the JSON output to extract only the detected fact error type and reasoning for each sentence. 

%Ths statistics of our training data is described in {\color{red}Appendix X}. 
In Appendix, Table \ref{tab:label-data-comparison}  contrasts our dataset with the aggregated data with human feedback.
The example of user and assistant prompts used for fine-tuning are provided in Table \ref{tab:prompt-example}.

\begin{table*}
\centering
\begin{small} 
\newcolumntype{Y}{>{\centering\arraybackslash}X}
\begin{tabularx}{\textwidth}{l| Y Y Y Y Y Y Y |Y}
\toprule
Error Category & OutE & EntE & PredE & CirE & GramE & LinkE & CorefE & Mean \\
\midrule
Random Guessing & 14.3\% & 14.3\% & 14.3\% & 14.3\% & 14.3\% & 14.3\% & 14.3\% & 14.3\% \\
\midrule
Zero-shot & 9.2\% & 29.0\% & 11.9\% & 7.6\% & \textbf{24.0}\% & 0.0\% & \textbf{18.2}\% & 14.3\% \\
\midrule
\textbf{LLM Feedback (Ours)} & \textbf{28.5}\% & \textbf{52.5}\% & \textbf{40.5}\% & \textbf{30.9}\% & 22.2\% & \textbf{20.0}\% & 0.0\% & \textbf{27.8}\% \\
\bottomrule
\end{tabularx}
\end{small}
\vspace*{-0.3cm}
\caption{Factuality error localization on 7 error categories. ``Zero-shot" is the results of Llama-3-8B-Instruct with zero-shot inference, while ``LLM Feedback" is Llama-3-8B-Instruct fine-tuned on LLM feedback. "Random Guessing" is the performance of randomly selecting from the seven categories, i.e., 1/7=14.3\%.}
\vspace*{-0.5cm}
\label{tab:main-localization-exp}
\end{table*}
\vspace*{-0.05cm}
\section{Evaluation}
\vspace*{-0.05cm}

\noindent\textbf{Methods.} We compare our fine-tuned model with several counterparts: (1) QA- and NLI-based methods, including \emph{QAFactEval}~\cite{fabbri2022qafacteval} and \emph{SummaC}~\cite{laban2022summac}; (2) Llama-3-8B-Instruct with \emph{zero-shot} inference with FineSurE's prompt; (3) fine-tuned with \emph{human feedback}. Contrary to (1) and (3), our model is only exposed to fine-grained LLM-generated feedback. In addition, for (3), it is not possible to localize error types due to the lack of available human error types annotated.

\smallskip\smallskip
\noindent\textbf{Metrics.} We follow the widely used metrics in recent works\,\cite{song2024finesure, liu2023improving}, verifying the agreement with human in three different levels: \emph{balanced accuracy} (bAcc), an indicator of sentence-level verification accuracy; \emph{summary-level} correlation, an indicator of agreement with humans' summary-level scores; \emph{system-level} correlation, an indicator of agreement with humans' ranking across different summarizers. Detailed description is provided in Appendix \ref{sec:Metrics}.

\vspace*{-0.05cm}
\subsection{Agreement with Humans}
\vspace*{-0.05cm}
%Our experimental results indicate that fine-tuning a Llama-3-8B-Instruct model using a large-scale LLM-labeled dataset outperforms using human-labeled datasets. Notably, the model's performance is significantly improved when fine-tuned with a combination of fact-check labels, reasoning, and error localization, compared to zero-shot performance. Additionally, both the model's error localization ability and overall performance increase as the proportion of LLM data grow.

Table \ref{tab:main-result-exp} shows the agreement with human judgment on test datasets, as described in Section \ref{sec:test-desc} of Appendix.  
\textbf{Training with a large amount of LLM-generated feedback outperforms using a limited set of human feedback}.
Although fine-tuning on humans' binary feedback exhibits higher agreement than solely relying on zero-shot inference, the improvement achieved via LLM feedback is much greater due to the ease of acquiring a larger volume of feedback. In addition, even when compared with previous QA- and NLI-based evaluators, our model maintains its dominance at all levels of evaluation. The analysis per data domain is detailed in Appendix \ref{sec:exp_per_domain}.

\begin{table}[t!]
\centering
\footnotesize
\newcolumntype{Y}{>{\centering\arraybackslash}X}
\begin{tabularx}{\columnwidth}{l |l |Y Y Y}
\toprule
\multirow{4}{*}[+0.1em]{\rotatebox[origin=c]{90}{\makecell{Llama-3-8B-Inst.\\(fine-tuned)}}} & \multirow{2}{*}{Setting} & Sent. & Summ. & Sys. \\
\cmidrule(lr){3-3} \cmidrule(lr){4-4} \cmidrule(lr){5-5}
& & bAcc & Pearson & Rank \\
\cmidrule(lr){2-5}
& Binary Label & 73.0\% & 0.628 & 0.649 \\
& + Reasoning & 71.9\% & 0.628 & 0.825 \\
& + Error Localization & 73.4\% & 0.625 & 0.865 \\
\bottomrule
\end{tabularx}
\vspace*{-0.3cm}
\caption{Ablation on the granularity of LLM feedback.}
\label{table:granularity-feedback}
\vspace*{-0.5cm}
\end{table}

\subsection{Factuality Error Localization}

Another advantage of using LLM-based feedback is its fine granularity, which allows for the specification of even factuality error types. Table \ref{tab:main-localization-exp} presents the accuracy of error localization across seven categories. Despite the 57.4\% of bAcc achieved by zero-shot inference, it only achieves very low performance in localization, which is almost the same as the mean accuracy of random guessing. However, when fine-tuned with LLM feedback, the mean accuracy improves from 14.3\% to 27.8\%\footnote{The 0.0\% of CorefE is due to its rarity in training data. We leave this issue as future work.}. Therefore, \textbf{fine-tuning with LLM feedback enhances the error localization capability over zero-shot inference.} 

%The zero-shot model demonstrated varied performance, achieving the highest accuracy in EntE (29.0\%) and the lowest in LinkE (0\%), with an overall mean accuracy of 14.3\%. In contrast, the fine-tuned model (\lowercase\expandafter{\romannumeral3}) significantly improves accuracy across almost all categories, particularly in EntE (52.5\%), PredE (40.5\%), and CirE (30.9\%), achieving an overall mean accuracy of 27.8\%.

%These results underscore the effectiveness of fine-tuning with a combination of fact-check labels, reasoning, and error localization. This approach enhances the model's ability to localize error categories. The fine-tuned model outperforms both the zero-shot model and the random guessing baseline. This highlights the value of targeted training in improving model performance.

\subsection{Ablation on Feedback Granularity}

We adjust the granularity of LLM feedback in three ways: (1) using only the binary labels indicating whether each sentence is factually correct or not (see Figure \ref{fig:ablation-binary}); (2) adding a reasoning step like the chain-of-thought in prompt engineering (see Figure \ref{fig:ablation-reasoning}); and (3) transforming the task to error localization (see Figure \ref{fig:finesure-prompt}). Table \ref{table:granularity-feedback} shows the change in agreement with humans as we add more information to LLM feedback. 

Solely relying on binary feedback exhibits fairly high bAcc but results in the lowest system-level correlation with humans. The addition of reasoning slightly decreases bAcc but improves system-level correlation. Further addition of error categorization synergizes with the reasoning addition, resulting in the best bAcc and system-level correlation. Therefore, \textbf{adding more explainable information to LLM feedback in fine-tuning results in better agreement with humans}.

\vspace*{-0.05cm}
\subsection{Ablation on Feedback Size}
\vspace*{-0.05cm}

To value the effectiveness of LLM feedback, we ablate the size of training data in fine-tuning, as summarized in Table \ref{table:data-size}. 25.0\% of our training data (25,660 LLM feedback) ensures better agreement than using 5,853 human feedback in fine-tuning. This explains that 5 LLM feedback are likely worth 1 human feedback. Moreover, \textbf{increasing the volume of training data with LLM feedback shows almost continuous improvement in fact verification performance}.

% 25% 이상 써야함 => 개수로 환산하면 25,660 >> 5,853, 12.5%: 12,830

%In our experiments, we observed a trend indicating that performance improves as the size of the LLM-labeled dataset increases. Our experimental setting involved starting without fine-tuning (0\%) and then fine-tuning the model with LLM-labeled data, beginning at 12.5\% and doubling each time until reaching 100\%. This step-wise increase allowed us to systematically assess the impact of dataset size on performance. This result provides the evidence of the benefits of utilizing larger LLM-labeled datasets.

%As shown in Table 3, bAcc for sentence classification improves from 57.4\%  to 73.4\% gradually. Similarly, the Pearson correlation for summary-level correlation shows a significant increase from 0.246 to 0.625. System-level rank correlation also follows this trend, improving from 0.663 to 0.865.

\begin{table}
\centering
\footnotesize
\newcolumntype{Y}{>{\centering\arraybackslash}X}
\begin{tabularx}{\columnwidth}{l |r |Y Y Y}
\toprule
\multirow{6}{*}[-1em]{\rotatebox[origin=c]{90}{\makecell{Llama-3-8B-Inst.\\(fine-tuned)}}} & \multirow{2}{*}{Setting} & Sent. & Summ. & Sys. \\
\cmidrule(lr){3-3} \cmidrule(lr){4-4} \cmidrule(lr){5-5}
& & bAcc & Pearson & Rank \\
\cmidrule(lr){2-5}
& 100.0\% & 73.4\% & 0.625 & 0.865 \\
& 50.0\% & 69.4\% & 0.601 & 0.902 \\
& 25.0\% & 71.6\% & 0.588 & 0.787 \\
& 12.5\% & 68.6\% & 0.509 & 0.589 \\
& 0.0\% & 57.4\% & 0.246 & 0.663 \\
\bottomrule
\end{tabularx}
\vspace*{-0.3cm}
\caption{Ablation on the size of LLM feedback.}
\label{table:data-size}
\vspace*{-0.5cm}
\end{table}

\begin{table*}
\begin{small}
\centering
\setlength{\tabcolsep}{2pt}
\newcolumntype{Y}{>{\centering\arraybackslash}X}
\begin{tabularx}{\textwidth}{l |Y Y Y Y}
\toprule
& Llama-3-8B-Inst.\hspace{-2em} & Llama-3-70B-Inst. & ChatGPT-3.5-Turbo & ChatGPT-4-Turbo \\
& (fine-tuned)  & (zero-shot) & (zero-shot) & (zero-shot) \\
\midrule
bAcc & 73.4\% & 77.3\% & 64.0\% & 79.3\%
\\
\midrule
Inference Time & 4.948s & 15.761s & 1.682s* & 8.462s* \\
\midrule
API Cost & 0\$ & 0\$ & 0.59\$ & 13.30\$ \\
\bottomrule
\end{tabularx}
\vspace*{-0.2cm}
\caption{Performance comparison with various LLMs. For inference, we used a batch size of 1 on a single NVIDIA H100 GPU. Quantization was applied to the 70B model to enable it to run on a single GPU. Inference time represents the time it takes for the LLM to generate a single piece of feedback. API cost refers to the expense incurred in generating feedback for 693 test examples. GPT series models used are \texttt{gpt-3.5-turbo-0125} and \texttt{gpt-4-turbo-2024-04-09}. * indicates response time.}
\vspace*{-0.3cm}
\label{tab:latency}
\end{small}
\end{table*}

\subsection{Inference Latency}
\label{sec:latency-analysis}
Table \ref{tab:latency} shows that \textbf{our fine-tuned model is more cost- and computing-efficient than other LLMs while keeping high performance.} From the perspective of knowledge distillation, our model achieved performance close to 95\% of the teacher model, Llama-3-70B-Instruct, while delivering over 3x faster inference time. Furthermore, when compared to the more affordable commercial model, ChatGPT-3.5-Turbo, our model exhibited significantly better performance. It also achieved approximately 1.7x faster inference time than ChatGPT-4-Turbo, along with substantial advantages in terms of API cost.

\subsection{Understanding Why It Works}
In this section, we discuss why training with LLM-generated feedback outperforms human feedback. Human evaluation becomes unreliable when summary feedback is fine-grained, such as identifying error types or providing explainable reasons. In the Appendix, Table \ref{tab:metadata-stat} shows that existing fine-grained human-labeled datasets have an inter-annotator agreement (Kappa) below 0.5, indicating low reliability of human labels. Therefore, the quality difference between LLM-generated labels and human labels is not significant. Based on this observation, according to the scaling law for LLM \cite{kaplan2020scaling}, an increase in the amount of training data is expected to enhance the performance of our model.
\vspace*{-0.1cm}
\section{Conclusion}
\vspace*{-0.1cm}

We release FineSumFact, a large-scale training dataset with LLM feedback, which can be used to train a fact verification model. We test multiple strategies to fine-tune LLMs w.r.t the granularity and the size of LLM feedback. The results indicate that fine-tuning with LLM feedback has the potential to create an effective and efficient fact verifier, addressing the lack of human feedback in training automated fact verification models.

%We introduce LLM-labeled datasets for fine-tuning LLMs for summarization tasks. For comparison, we also aggregate existing human-labeled datasets. We fine-tuned Llama-3-8B-Instruct model using human-labeled datasets and LLM-labeled datasets respectively. The results indicate that Fine-tuning automatic fact verification models with LLM-label are more effective and cost-efficient than using human labels. Therefore, our research verifies that LLM labels can overcome the limitations of human labels in automatic summary evaluation model training.

\section*{Limitations}
We report two main limitations in our study.

Firstly, summary feedback was generated from a single model, Llama-3-70B-Instruct. Therefore, we are unable to reflect feedback from diverse distributions. If we generate feedback using various LLMs, we would be able to generate more accurate feedback. 
Additionally, our training model, Llama-3-8B-Instruct, is fine-tuned using data comprised of summaries generated by 10 LLMs and feedback generated by Llama-3-70B-Instruct. Consequently, from a knowledge distillation perspective, the performance of the fine-tuned model may not surpass that of the LLMs used to generate the LLM feedback. 

%Secondly, the quality of LLM feedbacks may not always align perfectly with human judgment. Human-labeled data has been regarded as the most reliable ground truth, and LLM feedback has been less commonly used for model training due to potential discrepancies in judgment and occasional issues such as hallucinations. As LLMs continue to advance, particularly in mitigating hallucination issues, their outputs are expected to more closely align with human judgment.

Secondly, as discussed in Section 5.2, our dataset with LLM feedback presents some error category imbalance. Despite generating summaries using 10 LLMs, there was a lack of diversity in terms of error types. In the generated summary, there is significant inclusion of out-of-context error (OutE) and entity error (EntE), while coreference error (CorefE) is notably less frequent. Therefore, it was challenging to analyze performance by error type in error localization. Generating summaries synthetically to include a variety of error types could be a solution.

These challenges remain as future work. 

\section*{Ethics Statement}
There are no significant ethical concerns related to this work. Since there is no procedure involving human participation, there are no issues of bias. Additionally, we followed the copyright regulations, and there are no related concerns.

\section*{Acknowledgments}
This work was supported by the National Research Foundation of Korea (NRF) grant funded by the Korea government (MSIT) (No. RS-2024-00334343) and, additionally, supported by Institute of Information \& communications Technology Planning \& Evaluation (IITP) grant funded by the Korea government (MSIT) (No. RS-2024-00445087, Enhancing AI Model Reliability Through Domain-Specific Automated Value Alignment Assessment).

%\bibliography{jihwan/anthology,custom_new}

\appendix
\clearpage
\appendix
% -------- Appendix A----------
\begin{table*}[!t]
\vspace*{-0.4cm}
\footnotesize
\begin{center}
\begin{tabularx}{\textwidth}{l |l |l l X}
\toprule
\textbf{Dataset} & \textbf{Source} & \textbf{Annotators} & \textbf{Kappa} & \textbf{Annotation Scheme} \\
\midrule
\makecell[l]{AggreFact ${\dagger}$ \\ \small \cite{tang2022understanding}} & \makecell[l]{CNN/DM \\ XSUM} & \makecell[l]{mixed} & \makecell[l]{-} & \makecell[l]{binary} \\
\midrule
\makecell[l]{DiaSumFact \\ \small \cite{zhu2023annotating}} & \makecell[l]{SAMSum \\ QMSum} & 
\makecell[l]{2 in-house students} & \makecell[l]{0.49} & 
\small \makecell[X]{\{NoE, EntE, PredE, CirE, CorefE, LinkE, Others\}} \\
\midrule
\makecell[l]{TofuEval \\  \cite{tang2024tofueval}} & 
\makecell[l]{MediaSum \\ MeetingBank} & \makecell[l]{2 expert linguists} & \makecell[l]{0.42 \\ 0.34} & 
\makecell[X]{binary / \{extrinsic information error, misreferencing error, stating opinion as fact error, reasoning error, tense/modality error, contradiction error, nuanced meaning shift error, others\}} \\
\midrule
\makecell[l]{Ramprasad'24 \\  \cite{ramprasad2024evaluating}} & \makecell[l]{BillSum \\ PubMed} & \makecell[l]{2 expert attorneys \\ 2 expert medical doctors} & 
\makecell[l]{0.17 \\ 0.11} & 
\makecell[X]{binary / \{intrinsic, extrinsic, mixed, others\}} \\
\bottomrule
\end{tabularx}
\end{center}
\vspace*{-0.4cm}
\caption{Summary of the human-labeled datasets. We report Cohen's kappa in the original by default. In DiaSumFact, we report the average Cohen's kappa across six annotation groups, each consisting of two annotators. ${\dagger}$: we do not report Cohen's kappa since AggreFact integrates various datasets, some of which include Cohen's kappa values and others that do not.}
\label{tab:metadata-stat}
\vspace*{-0.0cm}
\end{table*}

\vspace{1cm}

\begin{table*}
\centering
\begin{small}
\renewcommand{\arraystretch}{1}
\newcolumntype{Y}{>{\centering\arraybackslash}X}
\begin{tabularx}{\textwidth}{l |Y Y Y Y >{\centering\arraybackslash}p{6em}}
\toprule
& Number of Documents & Number of Summaries & Number of Summarizers & Number of Domains & Doc. Length in Words \\
\midrule
Data with Human Feedback & 2,499 & 5,853 & 17 & 6 & 81-2,989 (531) \\
Data with LLM Feedback (Ours) & 10,877 & 102,640 & 10 & 7 & 5-3,847 (910) \\
\bottomrule
\end{tabularx}
\end{small}
\vspace*{-0.2cm}
\caption{Comparison of training datasets with human and LLM feedback. Doc. Length in Words indicates the min-max (average) of the document length in words.}
\label{tab:label-data-comparison}
\vspace*{-0.2cm}
\end{table*}

\vspace{1cm}

\begin{table*}[t]
\centering
\renewcommand{\arraystretch}{1.2}
\footnotesize
\begin{tabular*}{\textwidth}{@{\extracolsep{\fill}}l|ll|ll|ll|cc}
\toprule
\multirow{2}{*}{\textbf{Source Dataset}} & \multicolumn{2}{c}{\textbf{\# of Doc.}} & \multicolumn{2}{c}{\makecell{\textbf{\# of Label 0}}} & \multicolumn{2}{c}{\makecell{\textbf{\# of Label 1}}} & \multicolumn{2}{c}{\makecell{\textbf{Doc. Length in Words}}} \\ \cmidrule(lr){2-9}
 & \textbf{Train} & \textbf{Test} & \textbf{Train} & \textbf{Test} & \textbf{Train} & \textbf{Test} & \textbf{Train} & \textbf{Test} \\ \midrule
AggreFact       & 4,130 & 111 & 2,325 & 41 & 2,754 & 83  & 81-2,147 (470.5) & 165-1,303 (501.4) \\
DiaSumFact      & 339  & 27 & 599  & 43 & 394 & 28 & 93-585 (274.1) & 97-342 (238.7)   \\ 
TofuEval        & 1,241 & 531 & 2,751 & 1,167 & 641 & 322 & 710-1,199 (963.7) & 739-1,185 (919.3)   \\ 
Ramprasad'24   & 143  & 24 & 421  & 65 & 19  & 6 & 682-2,989 (1,725.9) & 916-2,349 (1,721.9)   \\ 
\bottomrule
\end{tabular*}
\vspace*{-0.3cm}
\caption{Statistics of train and test data annotated by humans according to the source. The number of documents (\# of Docs) is counted at the document level, while the number of labels (\# of Labels) is counted at the sentence level. The 'Doc. Length in Words column indicates 'min-max (average)' of the document length in words.}
\label{table:train_test_statistics}
\vspace*{-0.3cm}
\end{table*}

\section{Human-labeled Dataset Details}
\label{sec:human-labeled-data}

Table \ref{tab:metadata-stat} summarizes the details of the annotations for each dataset in our human-labeled datasets. The aggregation of these datasets covers various domains and text types. We briefly describe the human-labeled dataset we used. 

\subsection{Source Datasets}
\smallskip\smallskip
\noindent\textbf{AggreFact} \cite{tang2022understanding} is a factuality evaluation bencmark that includes two datasets in the news domain; CNN/DM \cite{hermann2015teaching} and XSum \cite{narayan2018don}. AggreFact integrates nine datasets from FactCC \cite{kryscinski2020evaluating}, Wang'20 \cite{wang2020asking}, SummEval \cite{fabbri2021summeval}, Polytope \cite{huang2020have}, Cao’22 \cite{cao2022hallucinated}, XSumFaith \cite{maynez2020faithfulness}, FRANK \cite{pagnoni2021understanding}, Goyal’21 \cite{goyal2021annotating}, and CLIFF \cite{cao2021cliff}. 
    
\smallskip\smallskip
\noindent\textbf{DiaSumFact} \cite{zhu2023annotating} collects fine-grained sentence-level factual error annotations for evaluating dialogue summarization. It spans two dialogue domains: daily conversations, containing chit-chat, and meetings, sourced from SAMSum \cite{gliwa2019samsum} and QMSum \cite{zhong2021qmsum}, respectively.
    
\smallskip\smallskip
\noindent\textbf{TofuEval} \cite{tang2024tofueval} contains two dialogue datasets for benchmarking automated evaluators in factuality. It covers two domains, Interview (MediaSum) \cite{zhu2021mediasum} and Meeting (MeetingBank) \cite{hu2023meetingbank}. Each summary is a topic-based summary generated by LLMs and includes sentence-level human annotations for factuality evaluation.

\smallskip\smallskip
\noindent\textbf{Ramprasad'24} \cite{ramprasad2024evaluating} addresses the news domain as well as two specialized domains: medicine (PubMed) \cite{cohan2018discourse} and legal (BillSum) \cite{kornilova2019billsum}. It releases human annotations from domain experts to assess the factuality of model-generated summaries.

\begin{table}[t!]
\centering
\footnotesize
\begin{tabularx}{\columnwidth}{l |r |r r r}
\toprule
\multirow{8}{*}[-0.2em]{\rotatebox[origin=c]{90}{\makecell{Llama-3-8B-Inst.\\\textbf{(zero-shot)}}}} & Error Type & Pred \#& Correct \# & Accuracy \\
\cmidrule(lr){2-5}
& OutE & 1,093 & 101 & 9.2\% \\
& EntE & 579 & 168 & 29.0\% \\
& PredE & 101 & 12 & 11.9\% \\
& CirE & 395 & 30 & 7.6\% \\
& GramE & 25 & 6 & 24.0\% \\
& LinkE & 9 & 0 & 0.0\% \\
& CorefE & 11 & 2 & 18.2\% \\
\bottomrule
\end{tabularx}
\vspace*{-0.3cm}
\caption{Error localization accuracy details in the zero-shot setting. Among the 9 error types, "No Error" and "Other Error" were excluded from the calculation. A total of 2,213 sentences were predicted as errors, of which 319 matched the correct error category.}
\label{}
\vspace*{-0.5cm}
\label{table:local-zero}
\end{table}

\subsection{Label Consolidation}
We aggregate human-labeled data at the sentence level from existing studies, which differ in terms of granularity and annotation schemes when labeling factual consistency. AggreFact and TofuEval provide majority-agreed binary labels for each summary sentence, eliminating the need for consolidation. However, Ramprasad'24 and DiaSumFact consist of two labels for each summary sentence from two annotators, without majority agreement. For these two datasets, if both annotators agreed on 'no error', the data was labeled as 'no error'. If they agreed on 'error' it was labeled as 'error'. If their results differed, the data was not included in our dataset.

\subsection{Dataset Split}
\label{sec:test-desc}
We aggregate a total of 6,546 document-summary pairs of human feedback data, consisting of 1,772 from TofuEval, 4,241 from AggreFact, 336 from DiaSumFact, and 167 from Ramprasad'24. Then, we split it into 5,853 training set and 693 test set. Data classified as the test set in the original datasets are also included as test data in our dataset. The test set is consistently used as ground truth to evaluate all the models in Table \ref{tab:main-result-exp}. Refer to Table \ref{tab:label-data-comparison} for a comparison between the entire set of human feedback data and LLM feedback data. The breakdown of human feedback data into train and test sets is summarized in Table \ref{table:train_test_statistics}.

\subsection{Testset for Error Localization}
We construct an additional test set of 1,286 document-summary pairs from FRANK\,\cite{pagnoni2021understanding}, which is a test set tailored for error localization evaluation in Table \ref{tab:main-localization-exp}. This dataset consists of labels annotated by three human annotators for each summary sentence across seven error types, identical to those in FineSurE\,\cite{song2024finesure}. In sentence-level error localization, if the model-predicted error type matched any one of the three human annotations, it was considered correct. Detailed error localization performance of Table \ref{tab:main-localization-exp} is shown in Tables \ref{table:local-zero} and \ref{table:local-fine}.

\begin{table}[t!]
\centering
\footnotesize
\begin{tabularx}{\columnwidth}{l |r |r r r}
\toprule
\multirow{8}{*}[-0.2em]{\rotatebox[origin=c]{90}{\makecell{Llama-3-8B-Inst.\\\textbf{(fine-tuned)}}}} & Error Type & Pred \#& Correct \# & Accuracy \\
\cmidrule(lr){2-5}
& OutE & 281 & 80 & 28.5\% \\
& EntE & 305 & 160 & 52.5\% \\
& PredE & 42 & 17 & 40.5\% \\
& CirE & 55 & 17 & 30.9\% \\
& GramE & 45 & 10 & 22.2\% \\
& LinkE & 40 & 8 & 20.0\% \\
& CorefE & 1 & 0 & 0.0\% \\
\bottomrule
\end{tabularx}
\vspace*{-0.3cm}
\caption{Error localization accuracy details in the fine-tuning setting. Among the 9 error types, "No Error" and "Other Error" were excluded from the calculation. A total of 769 sentences were predicted as errors, of which 292 matched the correct error category.}
\label{table:local-fine}
\vspace*{-0.2cm}
\end{table}

% -------- Appendix B----------

\section{Fact Verification Prompts}
\label{sec:FineSurE_Prompts}

\begin{figure}[t!]
    \centering
    \includegraphics[width=\linewidth]{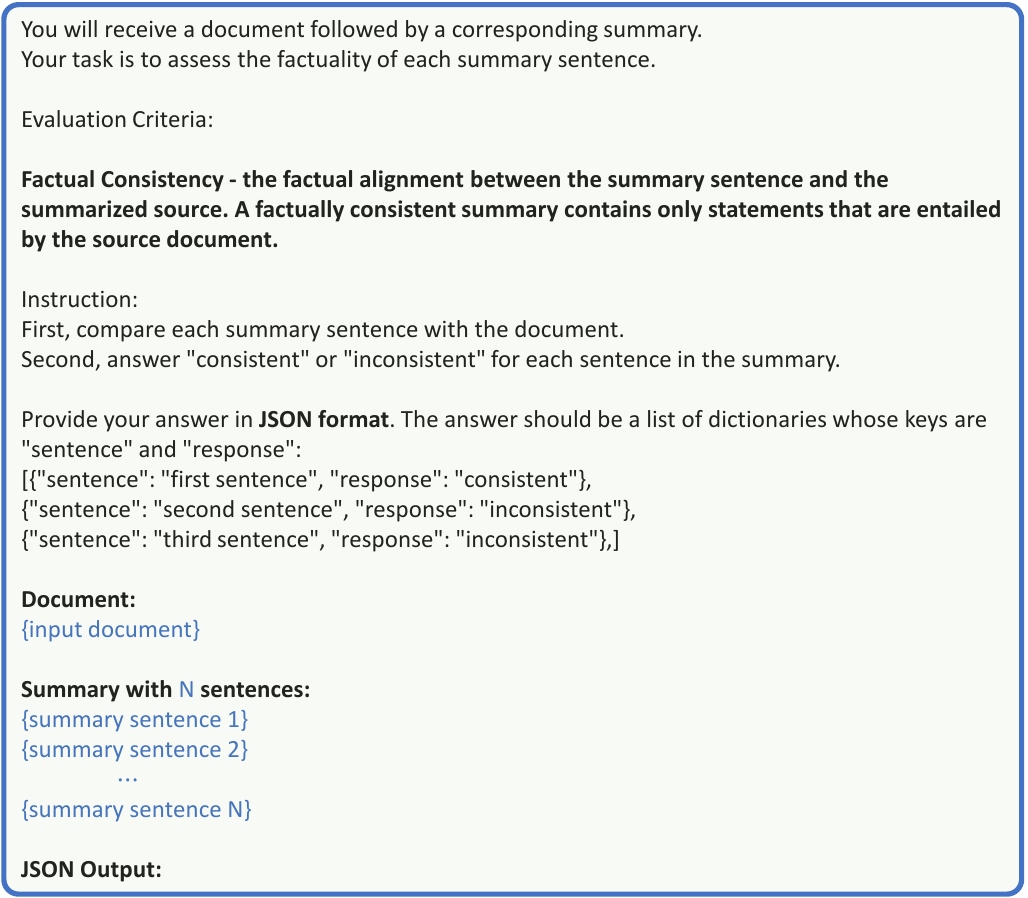}
    \vspace*{-0.7cm}
    \caption{Prompt for fact verification ("Binary Label" in Table \ref{table:granularity-feedback}).}
    \label{fig:ablation-binary}
    \vspace*{-0.1cm} % 첫 번째 그림과 두 번째 그림 사이의 간격 조정
\end{figure}

\begin{figure}[t!]
    \centering
    \includegraphics[width=\linewidth]{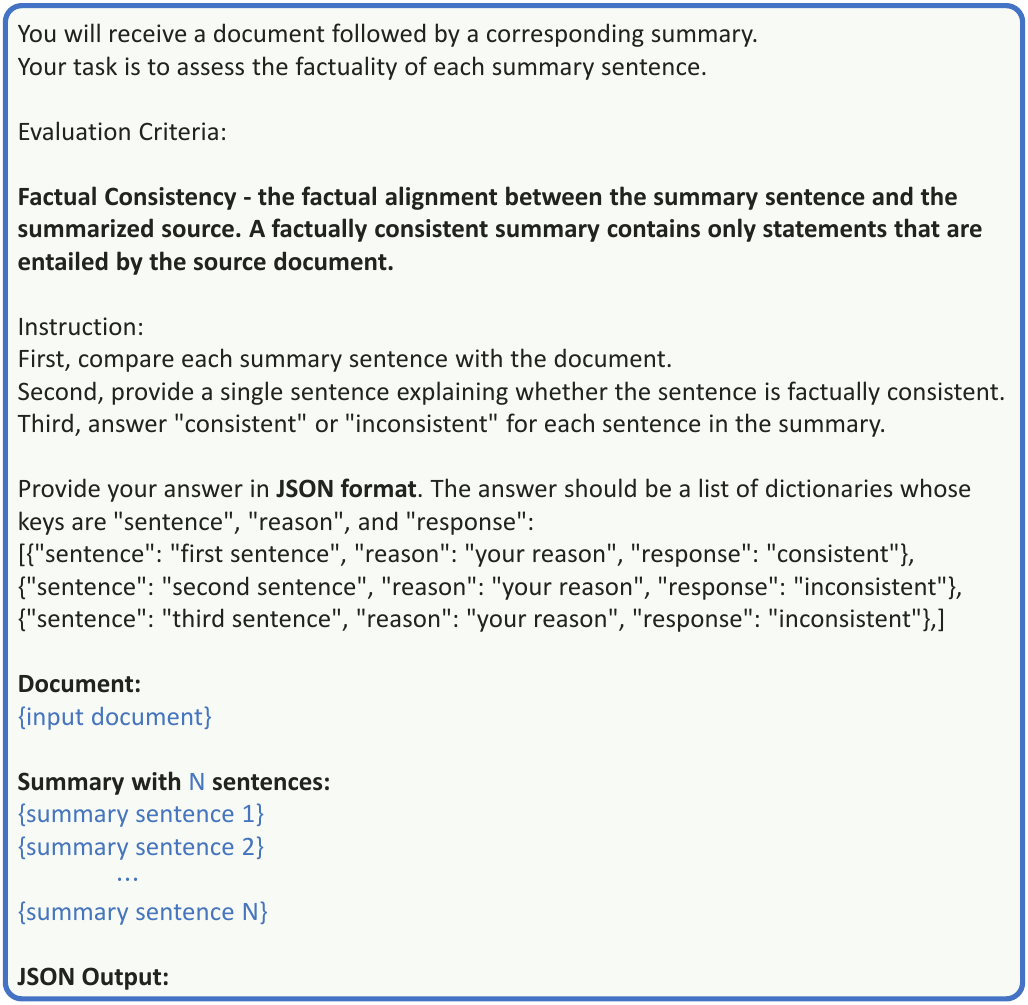}
    \vspace*{-0.7cm}
    \caption{Prompt for fact verification ("Binary Label + Reasoning" in Table \ref{table:granularity-feedback}).}
    \label{fig:ablation-reasoning}
    \vspace*{-0.3cm} % 두 번째 그림과 세 번째 그림 사이의 간격 조정
\end{figure}

\begin{figure}[t!]
    \centering
    \includegraphics[width=\linewidth]{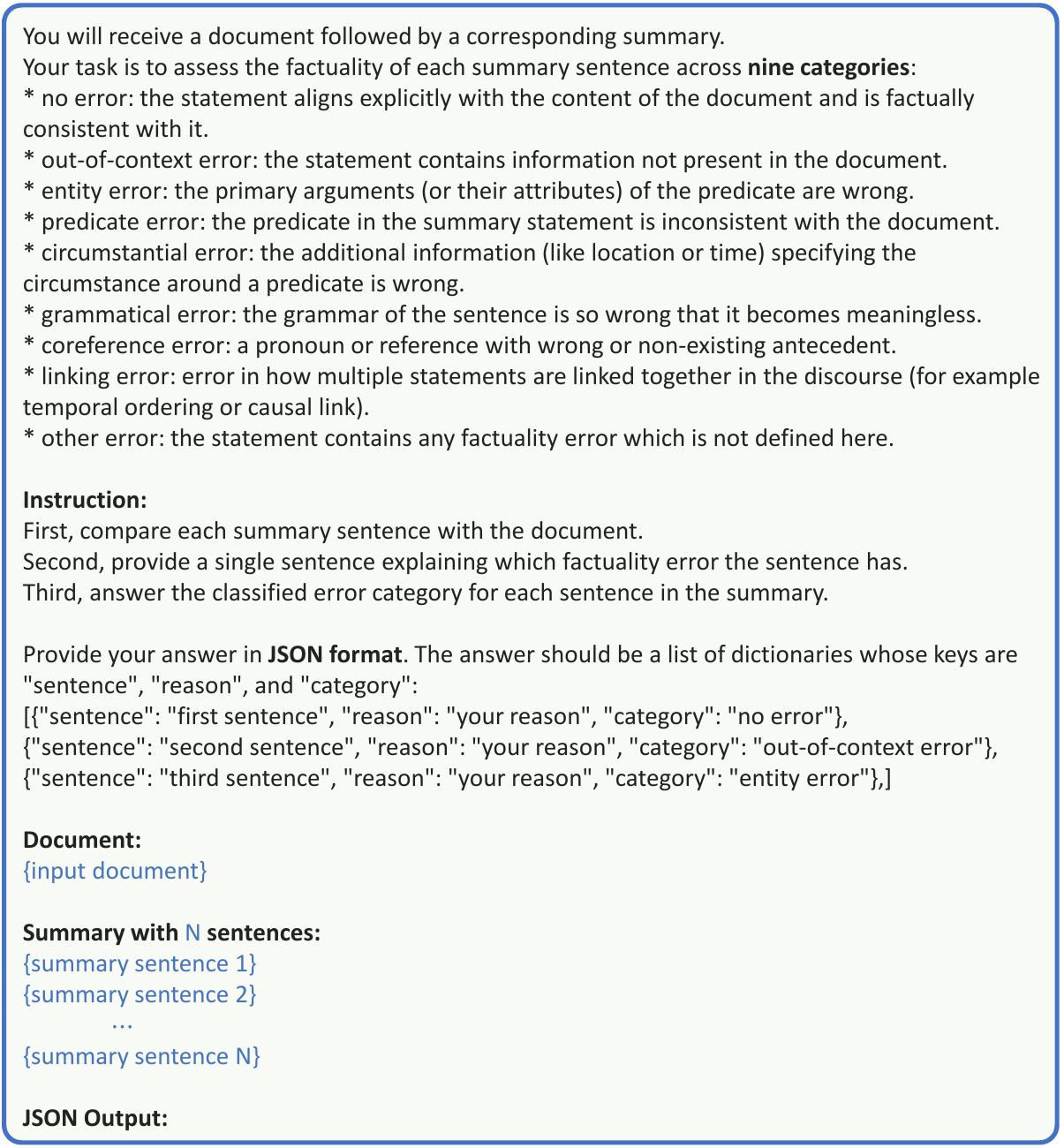}
    \vspace*{-0.7cm}
    \caption{Prompt for fact verification ("Binary Label + Reasoning + Error Localization" in Table \ref{table:granularity-feedback}, which is exactly the same with FineSurE \cite{song2024finesure}).}
    \label{fig:finesure-prompt}
    \vspace*{-0.3cm}
\end{figure}

We use three prompts to generate LLM feedback on fact verification, progressively increasing their granularity, as seen in our ablation of Table \ref{table:granularity-feedback}. 
The first one focuses on fact-checking using binary labels, as shown in Figure \ref{fig:ablation-binary}, while the second adds reasoning to the first one, as shown in Figure \ref{fig:ablation-reasoning}, and the third further incorporates error localization, as shown in Figure \ref{fig:finesure-prompt}.

Specifically, the first prompt asks the LLM to assess the factual consistency of each summary sentence against the source document, labeling sentences as either "consistent" or "inconsistent."
The second prompt adds complexity by requiring the LLM to not only judge consistency but also provide a brief explanation for each sentence's classification.
The third prompt further refines the process by asking the LLM to categorize specific types of factual errors across nine categories, allowing for detailed error identification.

\section{Factual Error Types}
\label{sec:error_types}
We follow the error taxonomy suggested by \citet{pagnoni2021understanding} for feedback generation. We provide explanations for each error category.

\smallskip\smallskip
\noindent\textbf{Out of Context Error (OutE)} indicate that summary statements include information not found in the document, which generally refers to an extrinsic error.

\smallskip\smallskip
\noindent\textbf{Entity Error (EntE)} means errors where the core arguments such as subject and object are wrong. This error typically occurs when the generated summary swaps entities.

\smallskip\smallskip
\noindent\textbf{Predicate Error (PredE)} refers to errors where the predicate in summary statements is not consistent with the document.

\smallskip\smallskip
\noindent\textbf{Circumstance Error (CirE)} occurs when additional information specifying the context around a predicate, such as location, time, or manner, is incorrect.

\smallskip\smallskip
\noindent\textbf{Grammatical Error (GramE)} encompasses errors in summary statements where significant grammatical mistakes make them meaningless.

\smallskip\smallskip
\noindent\textbf{Discourse Link Error (LinkE)} refers to error in how multiple statements are linked in the discourse, such as incorrect temporal ordering or causal links.

\smallskip\smallskip
\noindent\textbf{Coreference Error (CorefE)} is an error where pronouns or references are incorrect antecedents, causing ambiguity.

\begin{figure}[t!]
    \centering
    \includegraphics[width=\linewidth]{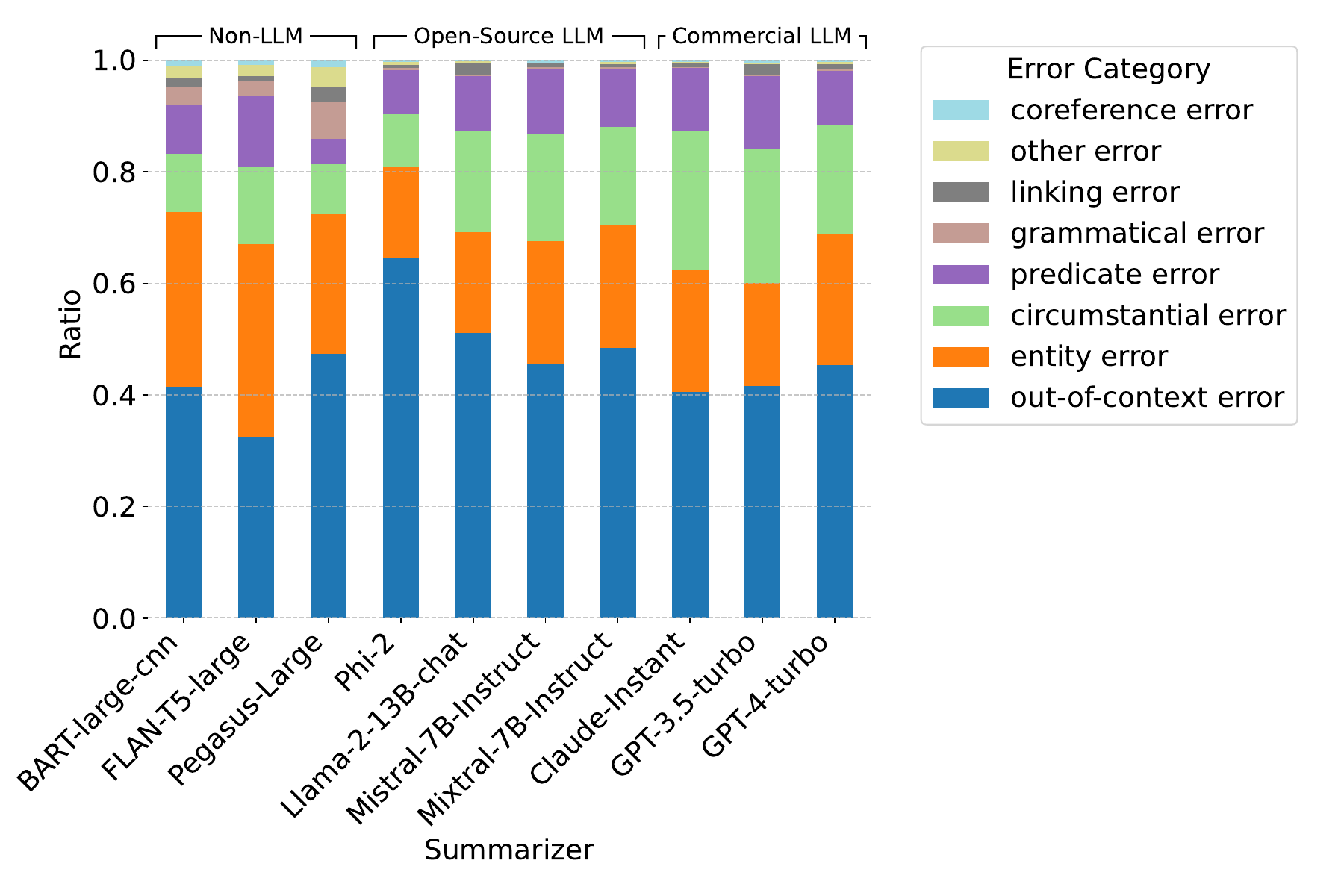}
    \vspace*{-0.7cm}
    \caption{Error category distribution of summaries with LLM feedback for each summarizer, where the error category is estimated using the automated fact verification.}
    \vspace*{-0.2cm}
    \label{fig:error_distribution}
\end{figure}

\begin{table*}[t]
\centering
\footnotesize
\begin{tabular*}{\textwidth}{@{\extracolsep{\fill}}llllc}
\toprule
\textbf{Source Dataset} & \textbf{\# of Doc.} & \textbf{\# of Label 0} & \textbf{\# of Label 1} & \textbf{Doc. Length in Words} \\ \midrule
CNN/DM          & 18,846 & 69,215 & 8,486 & 54-2,133 (774.8) \\ 
MediaSum     & 16,744 & 57,334 & 11,714 & 96-3,809 (1,546.3) \\ 
DialogSum    & 18,990 & 36,148 & 17,545 & 50-1,084 (187.6) \\ 
MeetingBank  & 14,427 &  41,187 & 14,169 & 126-3,847 (1,164.65)
\\
WikiHow    & 19,202 & 31,466 & 10,587 & 5-801 (83.7) 
\\
GovReport    & 3,045 & 14,378 & 728 & 180-3,777 (2,775.7)  \\ 
PubMed    & 11,386 & 47,757 & 3,606 & 21-3,721 (1,823.5) \\
\bottomrule
\end{tabular*}
\vspace*{-0.3cm}
\caption{Statistics of the training data labeled by LLM (Llama-3-70B-Inst.) according to the source. The number of documents (\# of Docs) is counted at the document level, while that of labels (\# of Labels) is counted at the sentence level. The 'Doc. Length in Words column indicates 'min-max (average)' of the document length in words.}
\vspace*{-0.5cm}
\label{table:machine_statistics}
\end{table*}

\begin{table}[ht]
\centering
\scriptsize
\begin{tabularx}{\columnwidth}{l |l}
\toprule
Model Name & HuggingFace/API Checkpoints \\
\midrule
\textbf{Summary Generation}\\
BART\textsubscript{large-cnn}& \texttt{facebook/bart-large-cnn} \\
FLAN-T5\textsubscript{large} & \texttt{google/flan-t5-large} \\
Pegasus\textsubscript{large} & \texttt{google/pegasus-large} \\
Phi-2 & \texttt{microsoft/phi-2} \\
Llama-2\textsubscript{13B-chat} & \texttt{meta-llama/Llama-2-13b-chat-hf} \\
Mistral\textsubscript{7B-Instruct} & \texttt{mistralai/Mistral-7B-Instruct-v0.1} \\
Mixtral\textsubscript{8x7B-Instruct} & \texttt{mistralai/Mixtral-8x7B-Instruct-v0.1} \\
Claude\textsubscript{Instant} & \texttt{claude-instant-1.2} \\
GPT-3.5\textsubscript{turbo} & \texttt{gpt-3.5-turbo-0125} \\
GPT-4\textsubscript{turbo} & \texttt{gpt-4-turbo-2024-04-09} \\
\midrule
\textbf{Feedback Generation}\\
Llama-3\textsubscript{70B-Instruct} & \texttt{meta-llama/Meta-Llama-3-70B-Instruct} \\
\midrule
\textbf{Fine-tuning}\\
Llama-3\textsubscript{8B-Instruct} & \texttt{meta-llama/Meta-Llama-3-8B-Instruct} \\
\bottomrule
\end{tabularx}
\vspace*{-0.2cm}
\caption{The model checkpoints.}
\label{tab:checkpoints}
\vspace*{-0.3cm}
\end{table}

\section{LLM Feedback Data Details}
\label{sec:llm-feedback-data-details}
\subsection{Dataset Construction} We generate 102,640 summaries from a total of 10,877 documents, including 18,846 from CNN/DM, 16,744 from MediaSum, 18,990 from DialogSum, 14,427 from MeetingBank, 19,202 from WikiHow, 3,045 from GovReport, and 11,386 from PubMed. The summaries and feedback are generated by 10 different language models and Llama-3-70B-Instruct, respectively. This dataset is exclusively used as a training set for fine-tuning the model, not as a test set. We provide the statistics of the LLM feedback dataset in Table \ref{table:machine_statistics}. Table \ref{tab:checkpoints} provides the details of the experiment's models. 

\subsection{Error Category Distribution}
We analyze the distribution of error categories based on feedback provided by Llama-3-70B-Instruct, evaluating summaries generated by 10 different LLMs. As shown in Table \ref{table:error_ratio}, the number of errors decreases as we move from non-LLMs to open-source LLMs and then to commercial LLMs. Additionally, we find that most summarizers exhibit a higher proportion of out-of-context errors and entity errors, while coreference errors are the least frequent. We provide the summary error type distribution in Figure \ref{fig:error_distribution}.

\begin{table}
\footnotesize
\centering
\newcolumntype{Y}{>{\centering\arraybackslash}X}
\begin{tabularx}{\columnwidth}{l |r r r}
\toprule
Summarizer & No Error & Error & Error Ratio \\
\midrule
\textbf{Non-LLM} & & & \\
BART\textsubscript{large-cnn} & 25,629 & 10,310 & 22.29\% \\
FLAN-T5\textsubscript{large} & 16,360 & 5,738 & 20.61\% \\
Pegasus\textsubscript{large} & 18,850 & 3,737 & 14.20\% \\
\midrule
\textbf{Open-Source LLM} & & & \\
Phi-2 & 10,917 & 11,559 & 33.96\% \\
Llama-2\textsubscript{13B-chat} & 20,040 & 4,618 & 15.77\% \\
Mistral\textsubscript{7B-Instruct} & 36,979 & 7,780 & 14.81\% \\
Mixtral\textsubscript{7B-Instruct} & 36,608 & 9,717 & 17.34\% \\
\midrule
\textbf{Commercial LLM} & & & \\
Claude\textsubscript{Instant} & 42,054 & 4,771 & 9.25\% \\
GPT-3.5\textsubscript{turbo} & 41,405 & 3,631 & 7.46\% \\
GPT-4\textsubscript{turbo} & 40,791 & 2,929 & 6.28\% \\
\bottomrule
\end{tabularx}
\vspace*{-0.2cm}
\caption{Error Ratio according to summarizes, indicates the proportion of summaries generated by each summarizer that are identified as errors by the feedback generator (Llama-3-70B-Instruct).}
\vspace*{-0.34cm}
\label{table:error_ratio}
\end{table}

\section{Agreement with Humans per Domain}
\label{sec:exp_per_domain}
As shown in Table \ref{tab:performance_domain_breakdown}, the performance across the News, Interview, and Meeting domains reveals varying levels of agreement with human judgments in fact verification, with each domain presenting unique challenges and insights.

\smallskip\smallskip
\paragraph{News} Both QA-based and LLM-based methods showed high agreement with human judgments, demonstrating their effectiveness in handling structured, fact-dense content typically found in news articles.

\smallskip\smallskip
\paragraph{Interview} LLM Feedback performed notably well, while QA-based and NLI-based methods struggled, underscoring the difficulties posed by the unstructured and conversational format of interview content.

\smallskip\smallskip
\paragraph{Meeting} The results were similar to those in the Interview domain, with the LLM Feedback method outperforming others. However, the overall performances of each type were lower, reflecting the inherent complexity in summarizing and verifying content from meetings.

\smallskip\smallskip
LLM-based methods stood out for their consistent performance across different domains. This robustness can be attributed to their fine-tuning with aggregated datasets that span a wide variety of domains, enabling them to generalize effectively across different types of content.

\begin{table*}[t]
\begin{center}
\footnotesize
\setlength{\tabcolsep}{12pt}
\begin{tabular}{c|c|cccccc}
\toprule
\multirow{2}{*}{Type} &
\multirow{2}{*}{Method} &
\multicolumn{2}{c}{News} &
\multicolumn{2}{c}{Interview} &
\multicolumn{2}{c}{Meeting} \\ 
\cmidrule(lr){3-4} \cmidrule(lr){5-6} \cmidrule(lr){7-8} 
& & 
Summ. & 
Sys.& 
Summ. & 
Sys. &
Summ. & 
Sys. \\ 
\midrule
QA-based                        
& QAFactEval &  0.614* & 0.886* & 0.406* & -0.257 & 0.382* & -0.167 \\
\midrule
\multirow{1}{*}{NLI-based} 
& SummaC-Conv & 0.515* & 0.683* & 0.208* & -0.086 & 0.168* & -0.433\\     
\midrule
\multirow{3}{*}{LLM-based}      
& Zero-shot  & 0.402* & 0.605* & 0.198* & 0.829* & 0.181* & 0.150\\
  
& Human Feedback & 0.287* & 0.560* & 0.443* & 0.600 & 0.468*  & 0.083\\
& LLM Feedback & 0.573* & 0.832* & 0.528* & 0.886* & 0.529* & 0.633\\
\bottomrule
\end{tabular}
\vspace*{-0.25cm}
\caption{Agreement with human judgments in fact verification on test data across three domains: News, Interview, and Meeting. The agreement was measured at the summary level using Pearson correlation (Summ.) and at the system level using rank correlation (Sys.) with a significance threshold of p-value < 0.05 (*). Results are reported per domain only when the test examples exceed 20. Domains with insufficient data, specifically \textbf{Daily}, \textbf{Legal}, and \textbf{Medicine} were excluded due to inflated p-values and statistically insignificant results.}
\label{tab:performance_domain_breakdown}
\end{center}
\vspace*{-0.3cm}
\end{table*}

\section{Metrics}
\label{sec:Metrics}
We follow the same settings as those presented in recent studies (\citealt{song2024finesure}, \citealt{liu2023improving}) to assess the model's performance and ensure alignment with human judgment.

\paragraph{bAcc}
\textit{Balanced accuracy (bAcc)} is used to address class imbalance when summarizing the performance of a model in a classification task. During sentence-level evaluation, human annotations and fine-tuned LLM classify factual correctness as '0' (No error) and incorrectness as '1' (Error). The formula for bAcc is as follows:

\begin{equation}
\begin{gathered}
\mathrm{bAcc} = \frac{\mathrm{TPR} + \mathrm{TNR}}{2}
\end{gathered}
\end{equation}

TPR(True Positive Rate), measures the proportion of correct positive predictions made by the fine-tuned LLM. TNR(True Negative Rate) quantifies the proportion of correct negative predictions made.

\paragraph{Faithfulness score}

For the summary-level and system-level evaluations, the percentage score of faithfulness enables us to assess summaries by aggregating sentence-level fact checks. Let us assume \( S_i = \{s_{i,1}, \ldots, s_{i,N}\} \) represents the \(i\)-th summary passage, consisting of \(N\) sentences, where \( s_{i,j} \) denotes the \(j\)-th sentence in the \(i\)-th summary passage. Additionally, let \( S_{i, \text{fact}} \subseteq S_i \) represent the subset of sentences identified as factually correct within this summary. The percentage score of faithfulness for \( S_i \), with respect to the original document \( D_i \), is computed as follows:

\begin{equation}
F(D_i, S_i) = \frac{|S_{i,\text{fact}}|}{|S_i|}
\end{equation}

\paragraph{Summary-level correlation}

To compute the \textit{summary-level correlation}, we define \( F_{gt} \) and \( F_{pred} \) as the faithfulness scores of the ground truth and the prediction, respectively. Let \( D = \{D_1, \ldots, D_k\} \) represent the set of input documents, and \( S = \{S_1, \ldots, S_k\} \) denote the corresponding set of summaries for these documents. Then the summary-level correlation is computed as follows:

\begin{equation}
\begin{gathered}
\mathrm{Pearson}([F_{gt}(D_{1}, S_{1}), \ldots, F_{gt}(D_{k},S_{k})], \\ [F_{pred}(D_{1}, S_{1}), \ldots, F_{pred}(D_{k}, S_{k})])
\end{gathered}
\end{equation}

\paragraph{System-level rank correlation}

To compute the \textit{system-level rank correlation}, we define \( \mathbf{F}_m = \{F_m(D_1,S_1), \ldots, F_m(D_k,S_k)\} \) as the set of percentage scores obtained from the labels given by the summarization model \( m \). Then, we construct a list of the average percentage scores for all summarization models, denoted as \( \left[\bar{\mathbf{F}}_{m_1}, \bar{\mathbf{F}}_{m_2}, \ldots \right] \) where, \( \bar{\mathbf{F}}_{m_i} = \frac{1}{|m_i|}\sum_{j=1}^{|m_i|}{F}_{m_i}(D_{j},S_{j})\). Using this list and the Rank function, we create the list \( \left[\mathrm{rank}_{m_1}, \mathrm{rank}_{m_2}, \ldots \right] \), where \( \mathrm{rank}_{m} \) represents the rank of model \( m \). By the same mechanism, we construct the ground truth list of ranks \( \left[\mathrm{rank}_{m_1}^*, \mathrm{rank}_{m_2}^*, \ldots \right] \) using the human labels. Finally, the summary-level correlation is computed as follows:

\begin{equation}
\begin{gathered}
\mathrm{Spearman}(\left[\mathrm{rank}_{m_1}, \mathrm{rank}_{m_2}, \dots\right], \\
\left[\mathrm{rank}_{m_1}^*, \mathrm{rank}_{m_2}^*, \dots\right]).
\end{gathered}
\end{equation}

The summary-level correlation indicates the agreement between human judgments and LLM, while the system-level rank correlation measures how closely the model rankings align with those provided by humans across various summarizers.

\begin{table*}[!htbp]
    \centering
    \raggedright{Document ID: PubMed-28493}\\
    \begin{tabularx}{\textwidth}{c| X}
        \toprule
        \multirow{2}{*}{\textbf{Input}} & \footnotesize \makecell[X]{You will receive a document followed by a corresponding summary. \\
        Your task is to assess the factuality of each summary sentence across \textbf{nine categories}: \\
        * no error: the statement aligns explicitly with the content of the document and is factually consistent with it. \\
        * out-of-context error: the statement contains information not present in the document. \\
        * entity error: the primary arguments (or their attributes) of the predicate are wrong. \\
        * predicate error: the predicate in the summary statement is inconsistent with the document. \\
        * circumstantial error: the additional information (like location or time) specifying the circumstance around a predicate is wrong. \\
        * grammatical error: the grammar of the sentence is so wrong that it becomes meaningless. \\
        * coreference error: a pronoun or reference with wrong or non-existing antecedent. \\
        * linking error: error in how multiple statements are linked together in the discourse (for example temporal ordering or causal link). \\
        * other error: the statement contains any factuality error which is not defined here. \\
        \\
        \textbf{Instruction:} \\
        First, compare each summary sentence with the document. \\
        Second, provide a single sentence explaining which factuality error the sentence has. \\
        Third, answer the classified error category for each sentence in the summary. \\
        \\
        Provide your answer in \textbf{JSON format}. The answer should be a list of dictionaries whose keys are \\
        "sentence", "reason", and "category": \\
        \textnormal{[\{"sentence": "first sentence", "reason": "your reason", "category": "no error"\}}, \\
        \textnormal{\{"sentence": "second sentence", "reason": "your reason", "category": "out-of-context error"\},} \\
        \textnormal{\{"sentence": "third sentence", "reason": "your reason", "category": "entity error"\},]} \\
        \\
        \textbf{Document:}\\
        Chronic condition management has been structured around models that aim at facilitating and development of care through initiatives of self - management , organisational restructuring and information system enhancement . uk s health care framework for chronic condition care draws upon the chronic care model and innovation care for chronic conditions framework . however , consideration to the clarity or vagueness of boundaries between different hierarchical structures and awareness of each structure about its focus and contribution to overall health system are not taken into account by these models . the proposed chronic condition management model is different as it is targeted at primary care which plays a salient role of gate keeping . change implemented here would lead to better clinical effectiveness , increase in efficiency of the system and reduction in cost . this model does not directly call for policy change ; instead it recommends incorporating available evidence into the policy . it calls for empowering the patient by providing health literacy and improving awareness , and for collaboration and cooperation between primary , secondary , tertiary , community and social care services . initiatives towards chronic condition management should be enacted at primary care due to its gate keeping role in the uk s health care system .\\ \\ \textbf{Summary with 3 sentences:}\\ \textnormal{[1]} The proposed chronic condition management model emphasizes the role of primary care in the UK's health care system for managing chronic conditions. \\
        \textnormal{[2]} This model advises incorporating available evidence into policies, empowering patients with health literacy and awareness, and fostering collaboration between primary, secondary, tertiary, community, and social care services. \\
        \textnormal{[3]} The model does not call for direct policy changes, but rather for addressing the clarity or vagueness of boundaries between hierarchical structures and making each structure aware of its focus and contribution to the overall health system. \\ \\
        \textbf{JSON Output:}} \\
        \midrule
        \multirow{1}{*}{\textbf{Output}} & \footnotesize \makecell[X]{[\\\{"sentence": "The proposed chronic condition management model emphasizes the role of primary care in the UK's health care system for managing chronic conditions.", "reason": "The sentence accurately summarizes the document's information about the proposed model's focus on primary care.", "category": "no error"\}, 
        \{"sentence": "This model advises incorporating available evidence into policies, empowering patients with health literacy and awareness, and fostering collaboration between primary, secondary, tertiary, community, and social care services.", "reason": "The sentence accurately summarizes the document's information about the model's recommendations.", "category": "no error"\}, \\
        \{"sentence": "The model does not call for direct policy changes, but rather for addressing the clarity or vagueness of boundaries between hierarchical structures and making each structure aware of its focus and contribution to the overall health system.", "reason": "The sentence incorrectly states that the model addresses the clarity or vagueness of boundaries between hierarchical structures, which is a criticism of existing models, not a feature of the proposed model.", "category": "out-of-context error"\},\\]
        }\\
        \bottomrule
    \end{tabularx}
    \vspace*{-0.2cm}
    \caption{An example of an input (user prompt) and its corresponding output (assistant prompt) used to fine-tune the Llama-3-8B-Instruct model.}
    \label{tab:prompt-example}
\end{table*}

\newpage

% \begin{table*}
% \centering
% \newcolumntype{Y}{>{\centering\arraybackslash}X}
% \begin{tabularx}{\textwidth}{l X X X X X X X X l}
% \toprule
% Summarizer & CirE & CorefE & EntE & GramE & LinkE & OterE & OutE & PredE & NoE \\
% \midrule
% BART-large-cnn & 1077 & 104 & 3238 & 327 & 188 & 211 & 4270 & 895 & 25629 \\
% Claude-Instant & 1184 & 14 & 1040 & 10 & 32 & 12 & 1937 & 542 & 42054 \\
% FLAN-T5-large-cnn & 797 & 49 & 1977 & 167 & 42 & 115 & 1868 & 723 & 16360 \\
% GPT-3.5-turbo & 872 & 17 & 668 & 8 & 69 & 10 & 1513 & 474 & 41405 \\
% GPT-4-turbo & 573 & 9 & 685 & 11 & 26 & 12 & 1329 & 284 & 40791 \\
% Llama-2-13B-chat & 835 & 5 & 834 & 11 & 101 & 15 & 2358 & 459 & 20040 \\
% Mistral-7B-Instruct & 1497 & 31 & 1702 & 24 & 50 & 16 & 3551 & 909 & 36979 \\
% Mixtral-7B-Instruct & 1714 & 36 & 2133 & 34 & 48 & 38 & 4707 & 1007 & 36608 \\
% Pegasus-Large & 335 & 47 & 933 & 250 & 96 & 132 & 1772 & 172 & 18850 \\
% Phi-2 & 1071 & 29 & 1897 & 50 & 60 & 71 & 7468 & 913 & 10917 \\
% \bottomrule
% \end{tabularx}
% \caption{Error codes: CE - Circumstantial Error, CoE - Coreference Error, EE - Entity Error, GE - Grammatical Error, LE - Linking Error, OE - Other Error, OCE - Out-of-Context Error, PE - Predicate Error. }
% \end{table*}

\end{document}